\documentclass[10pt,twocolumn,letterpaper]{article}

\usepackage[pagenumbers]{cvpr} 

\usepackage{graphicx}
\usepackage{amsmath}
\usepackage{amssymb}
\usepackage{booktabs}
\usepackage{multirow}
\usepackage[T1]{fontenc}
\usepackage[pagebackref,breaklinks,colorlinks]{hyperref}

\usepackage[capitalize]{cleveref}
\crefname{section}{Sec.}{Secs.}
\Crefname{section}{Section}{Sections}
\Crefname{table}{Table}{Tables}
\crefname{table}{Tab.}{Tabs.}

\begin{document}

\title{MSLKANet: A Multi-Scale Large Kernel Attention Network for Scene Text Removal}

\author{GuangtaoLyu\\
School of Computer Science and Artificial Intelligence, Wuhan University of Technology\\
China\\
{\tt\small 297903@whut.edu.cn}
}
\maketitle

\begin{abstract}
Scene text removal aims to remove the text and fill the regions with perceptually plausible background information in natural images. It has attracted increasing attention due to its various applications in privacy protection, scene text retrieval, and text editing. With the development of deep learning, the previous methods have achieved significant improvements. However, most of the existing methods seem to ignore the large perceptive fields and global information. The pioneer method can get significant improvements by only changing training data from the cropped image to the full image. In this paper, we present a single-stage multi-scale network MSLKANet for scene text removal in full images. For obtaining large perceptive fields and global information, we propose multi-scale large kernel attention (MSLKA) to obtain long-range dependencies between the text regions and the backgrounds at various granularity levels. Furthermore, we combine the large kernel decomposition mechanism and atrous spatial pyramid pooling to build a large kernel spatial pyramid pooling (LKSPP), which can perceive more valid pixels in the spatial dimension while maintaining large receptive fields and low cost of computation. Extensive experimental results indicate that the proposed method achieves state-of-the-art performance on both synthetic and real-world datasets and the effectiveness of the proposed components MSLKA and LKSPP. 

\end{abstract}

\section{Introduction}
\label{sec:intro}
Scene text images contain quite a lot of sensitive and private information, such as names, addresses, and cellphone numbers. With the increasing development of scene text detection and recognition technology, there is a high risk that the information collected automatically is used for illegal purposes. Therefore, a scene text removal task is proposed to solve this problem. Scene Text Removal aims at erasing text regions and filling the regions with perceptually plausible background information in natural images. It is also useful for privacy protection, image text editing \cite{wu2019editing}, video text editing \cite{subramanian2021strive}, and scene text retrieval \cite{wang2021scene}. 

\begin{table}
\center
  \caption{Train the pioneer method scene text eraser \cite{nakamura2017scene} on cropped image and full image.}
  \resizebox{\linewidth}{!}{\begin{tabular}{|cc|c|c|c|c|c|c|}
    \hline
    \multicolumn{2}{|c|}{Method}&PSNR&MSSIM&MSE&AGE&pEPs&pCEPs\\
    \hline
    \multicolumn{2}{|c|}{croped image} & 20.60 & 84.11 & 0.0233 & 14.4795 & 0.1304 & 0.0868\\
    \hline
    \multicolumn{2}{|c|}{full image} & 30.43 & 94.02 & 0.0018 & 5.2371 & 0.0301 & 0.0231 \\
    \hline
   \end{tabular}}
\label{tab_intro}
\end{table}

Scene text removal is a challenging task because it inherits the difficulty of both scene text detection and image inpainting tasks. With the development of deep learning, CNN-based \cite{nakamura2017scene,lyu2022psstrnet,wang2021pert} and GAN-Based methods \cite{zhang2019ensnet,liu2020erasenet,tursun2020mtrnet++} have achieved significant improvements. However, both scene text detection and image inpainting need global information, this problem seems to be ignored. As shown in table \ref{tab_intro}, we can get significant improvements by changing the cropped image to a full image, which also can demonstrate the importance of large perceptive fields and global information. Previous methods propose to predict the text stroke \cite{liu2020erasenet,wang2021pert} or combine the text region mask \cite{tursun2019mtrnet,tursun2020mtrnet++} as input to focus scene text removal only on these stroke regions to avoid the issue. In addition, many existing algorithms make few attempts to eliminate redundant feature information resulting in usually not better structure restoration and detail preservation while their receptive fields are limited and often cause inexhaustive erasure and unveracious inpainting due to lack of global information. Therefore, it is necessary to make full use of the text region features and the background features to explore the scale-space feature correlation from a global and local perspective. 

To address these issues, we propose a novel one-stage Multi-Scale Large Kernel Attention scene text removal network (MSLKANet) that combines multi-scale feature extraction\cite{qin2021multi}, large kernel attention \cite{guo2022visual}, large kernel Decomposition, and spatial pyramid pooling \cite{chen2017rethinking,he2015spatial}. Specifically, we propose a Multi-Scale Large Kernel Attention (MSLKA) block that combines multi-scale mechanism and large kernel attention to build various long-range correlations with relatively few computations. Moreover, to get larger receptive fields, we proposed a Large Kernel Spatial Pyramid Pooling(LKSPP) block, which was implemented by replacing the atrous convolution with decomposed large kernel convolution. More importantly, our proposed blocks are general in the sense that they could be easily applied to any network. 

We conducted extensive experiments on two benchmark datasets: SCUT-EnsText \cite{liu2020erasenet} and SCUT-syn \cite{zhang2019ensnet}. Both the qualitative and quantitative results demonstrate that MSLKANet can outperform previous state-of-the-art STR methods. The proposed MSLKA and LKSPP blocks can be used for other STR models for performance enhancement.

We summarize the contributions of this work as follows:
\begin{itemize}
\item We propose multi-scale large kernel attention (MSLKA) to obtain long-range dependencies between the text regions and the backgrounds at various granularity levels, therefore significantly improving the model representation capability.

\item We integrate the large kernel decomposition mechanism and atrous spatial pyramid pooling to build a large kernel spatial pyramid pooling (LKSPP), which can perceive more valid pixels in the spatial dimension while maintaining large receptive fields and low cost of computation.

\item Extensive experiments on both real and synthetic datasets demonstrate the performance of MSLKANet, significantly outperforming existing SOTA methods on quantitative and qualitative results. Ablation studies demonstrate the effectiveness of the proposed MSLKA and LKSPP blocks.
\end{itemize}

\begin{figure*}
  \includegraphics[width=\textwidth,height=0.4\textheight]{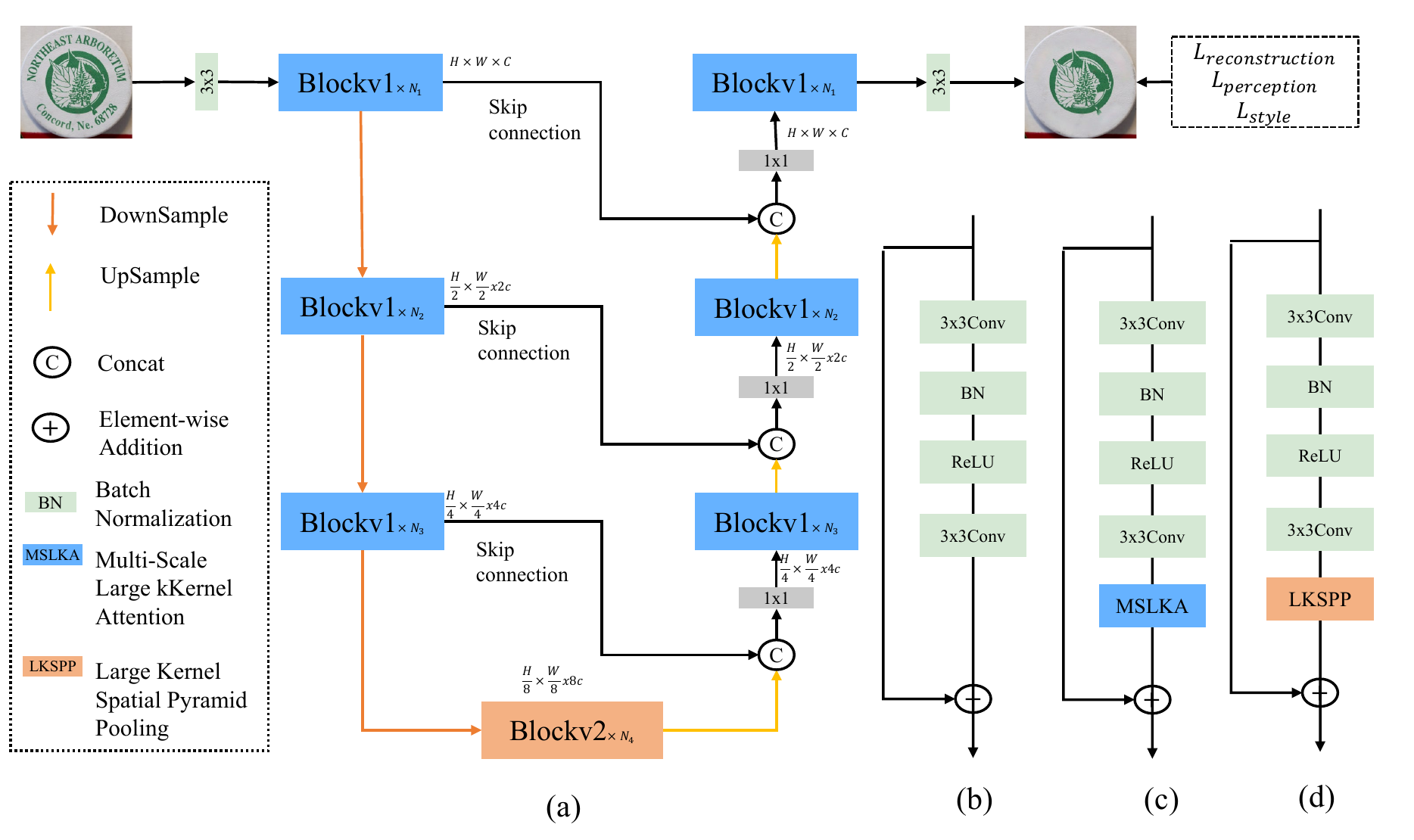}
  \caption{ (a) is the overall architecture of MSLKANet for scene text removal. It is adapted from the classic encoder-decoder-skip-connections architecture. (b) is the basic block used in the baseline. (c) and (d) is the basic block with proposed multi-scale large kernel attention (MSLKA) and large kernel spatial pyramid pooling (LKSPP).}
  \label{architecture}
\end{figure*}

\section{Related works}
Generally, the existing STR research can be classified into one-stage methods and two-stage methods. One-stage methods automatically detect the text regions and remove them in an end-to-end network. The method proposed in \cite{nakamura2017scene} is the first deep learning-based method to address the scene text removal task in an end-to-end way. It uses a single-stage U-shaped architecture with skip connections. However, their method only uses cropped image patches as input to train the network, which limits the equality of the scene text removal results due to the loss of global information. Scene Text removal can be taken as image-to-image translation. Inspired by Pix2pix \cite{isola2017image}, EnsNet \cite{zhang2019ensnet} proposed a refined loss and a local-aware discriminator to ensure the background reconstruction and integrity of the non-text region. EraseNet proposed an additional segmentation head to help locate the text and introduce a coarse-to-refinement architecture from image inpainting. MTRNet++ \cite{tursun2020mtrnet++} separately encoded the image content and text mask in two branches. Cho et al. \cite{cho2021detecting} proposed a two-task blending structure that performs both text segmentation and background restoration at the same time. PERT\cite{wang2021pert} progressively performed scene text erasure with a region-based modification strategy to only modify the pixels in the predicted text regions. PSSTRNet\cite{lyu2022psstrnet} proposed a novel mask update module to get a more and more accurate text mask for region-based modification strategy and an adaptive fusion strategy to make full use of results from different iterations.

The two-step methods decompose the text-removal task into two sub-problems: text detection and background in painting. MTRNet \cite{tursun2019mtrnet} concat the scene text image and text region mask as input. Zdenek et al. \cite{zdenek2020erasing} use the bounding boxes of text regions to train the text detection network and use the general natural scene images to train the inpainting network. This weak supervision method does not require paired training data. Conrad et al. \cite{conrad2021two} use a text detector to first get coarse text regions, and then use a segmentation network to refine the text pixels before the application of a pre-trainedEdgeConnect \cite{nazeri2019edgeconnect} for background inpainting. Bian et al. \cite{bian2022scene} proposed a cascaded GAN-based model, which decouples text stroke detection and stroke removal in text removal tasks.

\section{Method}
\subsection{Framework Overview}
Fig.\ref{architecture} shows the overall architecture of the proposed MSLKANet. It is based on the UNet-Style architecture with encoder, decoder, and skip-connections. Furthermore, we propose large kernel spatial pyramid pooling (LKSPP) and multi-scale large kernel attention (MSLKA) for obtaining long-range dependencies between the text regions and the backgrounds. In the loss functions, we use reconstruction loss, style loss \cite{gatys2016image}, and perception loss \cite{johnson2016perceptual} to keep the same with most existing methods.

\subsection{multi-scale large kernel attention}
\begin{figure}
  \includegraphics[width=\linewidth]{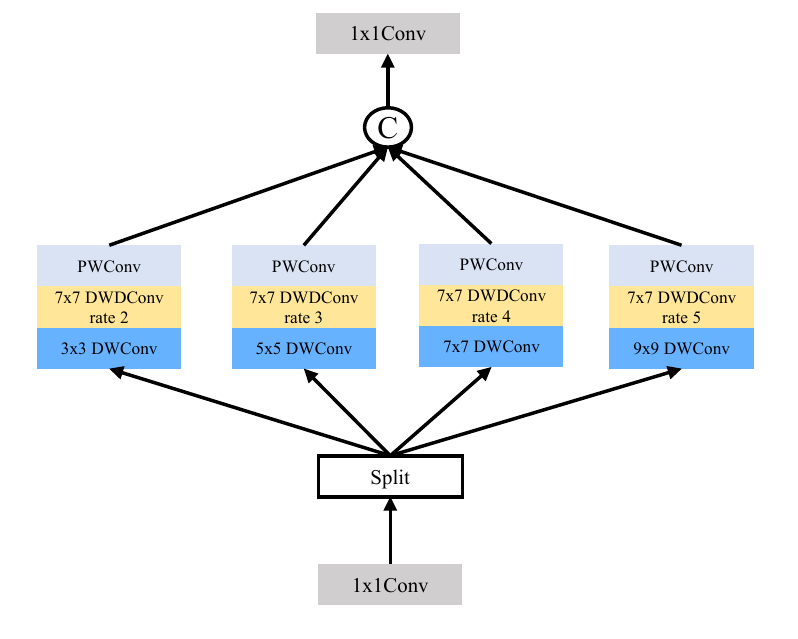}
  \caption{ Details of the proposed MSLKA. A $K \times K$  standard convolution into three components: a $\lceil \frac{K}{d} \rceil \times \lceil \frac{K}{d} \rceil$ depth-wise dilation convolution with dilation $d$, a $(2d-1) \times (2d-1)$ depth-wise convolution and a 1$\times$1 convolution.}
  \label{block_mslka}
\end{figure}

Attention mechanisms can help networks focus on important information and ignore irrelevant ones. And multi-scale feature fusion methods effectively combine image features at different scales and are now widely used to collect useful information about objects and their surroundings. However, they seem to be ignored by previous methods. Compare with other attention mechanisms(Channel Attention \cite{hu2018squeeze}, Self-Attention \cite{vaswani2017attention}), Large Kernel Attention \cite{guo2022visual} combines the advantages of convolution and self-attention. It takes the local contextual information, large receptive field, linear complexity, dynamic process, spatial dimension adaptability, and channel dimension adaptability into consideration. 

Under the guidance of the above ideas, we propose multi-scale large kernel attention (MSLKA) to combine large kernel decomposition and multi-scale learning for obtaining heterogeneous-scale correlations between the text regions and the background regions. 

\textbf{Large Kernel Attention.} LKA adaptively builds the long-range relationship by decomposing a $K \times K$ convolution into a $\lceil \frac{K}{d} \rceil \times \lceil \frac{K}{d} \rceil$ depth-wise dilation convolution with dilation $d$, a $(2d-1) \times (2d-1)$ depth-wise convolution and a 1$\times$1 convolution. Through the above decomposition, it can capture a long-range relationship with slight computational cost and parameters.

\textbf{Multi-Scale Large Kernel Attention}. To learn the attention maps with multi-scale longe range information, we improve the LKA with the group-wise multi-scale mechanism. As shown in Fig.\ref{block_mslka}, given the input feature maps $X \in  \mathbb{R}^{C \times H \times W} $, we split the input into different groups. For different groups, we use different dilation rates for various perceptive fields for multi-scale longe range information. In detail, we leverage four groups of LKA with a set of dilation rates$\{2,3,4,5\}$. By setting a fix-size kernel 5 in each depth-wise $d$-dilation convolution, we get a set of large kernels $\{10,15,20,25\}$. Through the above convolutions, we get the four scale attention maps and then reweight the different groups of features. In the last, we concat the four groups of features as output.

\subsection{large kernel spatial pyramid pooling}
\begin{figure}
  \includegraphics[width=\linewidth,height=0.8\linewidth]{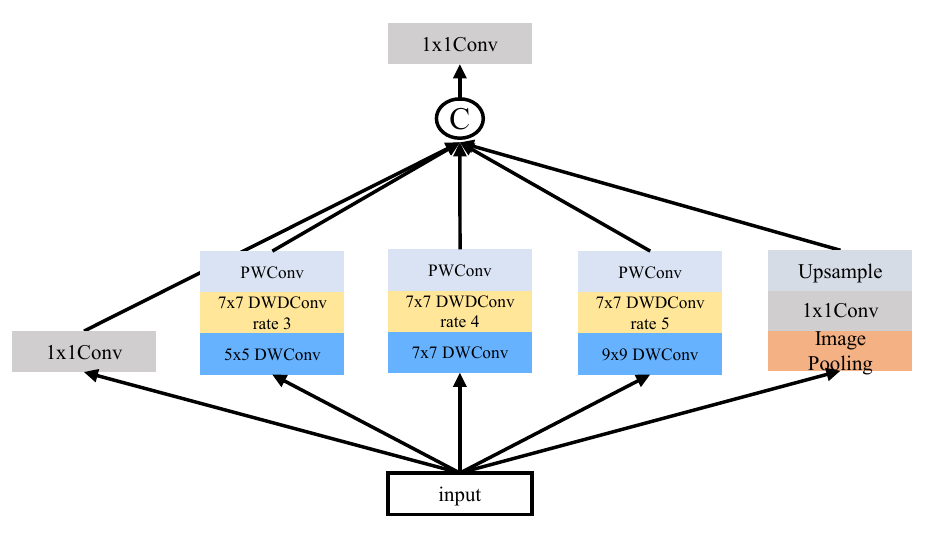}
  \caption{ Details of the proposed LKSPP. We use decomposed large kernel convolution to substitute the depthwise separable atrous convolution.}
  \label{block_lkaspp}
\end{figure}

For obtaining large perceptive field and multi-scale information, Spatial Pyramid Pooling(SPP) is another frequently-used method. Atrous Spatial Pyramid Pooling(ASPP) proposed and improved in \cite{chen2017rethinking,chen2017deeplab,chen2018encoder} is a most common and efficient block in different SPP methods(SPP\cite{he2015spatial}, PPM\cite{zhao2017pyramid}, ASPP). In Deeplabv2\cite{chen2018encoder}, the  ASPP block only includes several parallel atrous convolutions with different rates for getting multi-scale information and a 1x1 convolution. Image-level pooling ars insert into ASPP for obtaining global information in Deeplabv3 \cite{chen2017rethinking}. In Deeplabv3+\cite{chen2018encoder}, they use depthwise separable atrous convolution replacing the standard atrous convolution for reducing the cost of computation. Motivated by LKA and the development of ASPP, we use decomposed large kernel convolution to substitute the depthwise separable atrous convolution for the fusion of more valid pixels in the spatial dimension while maintaining large receptive fields and lower cost of computation. In detail, our large kernel spatial pyramid pooling (LKSPP) includes three parallel decomposed large kernel convolutions with different rates(3,4,5) and a fix-size kernel 7 in each depth-wise dilation convolution for getting multi-scale information, a 1x1 convolution, and an image-level pooling as shown in Fig.\ref{block_lkaspp}, . 

\subsection{Loss Functions}\label{sect3}

We introduce three loss functions for training MSLKANet, including reconstruction loss, perceptual loss, and style loss. Given the input scene text image $I_{i}$, text-removed ground truth (gt) $I_{gt}$, and the STR output from MSLKANet is denoted as $I_{o}$.

\textbf {Reconstruction Loss.} First of all, we apply $L_1$ distance is proposed to measure the output and the ground truth in the pixel level.

\begin{equation}
    L_{rec} = ||I_{o} - I_{gt}||,
    \label{eq_loss_rec}
\end{equation}

\textbf {Perceptual Loss.} To capture the high-level semantic difference and simulate the human perception of image quality, we utilize the perceptual loss\cite{johnson2016perceptual} in Eq.\eqref{eq_loss_per}. $\Phi_i$ is the activation features of the $i$-th layer of the VGG-16 backbone. 
\begin{equation}
\begin{split}
	L_{per} = ||\Phi_i(I_{o}) - \Phi_i(I_{gt})||_1
\end{split}
\label{eq_loss_per}
\end{equation}

\textbf {Style Loss.} We also use the style loss \cite{isola2017image} for reducing the global level style of output $I_{o}$ and its corresponding ground-truths $I_{gt}$ as Eq.\eqref{eq_loss_style}, where $G_i^(\Phi)$ is the Gram matrix constructed from the selected activation features in perceptual loss, and $|| ||_2$ is mean-squared distance.
\begin{equation}
 	L_{style} = ||G_i (\Phi(I_{o})) - G_i (\Phi(I_{gt}))||_2
\label{eq_loss_style}
\end{equation}

In Summary, the total loss for training MSLKANet is the weighted combination of the above mentioned losses:
\begin{equation}
\begin{split}
	L_{total} = L_{rec}+\lambda_s L_{style}+\lambda_p L_{per}
\end{split}
\end{equation}

In experiments, $\lambda_s$, $\lambda_{p}$ is set to be 120,0.01, respectively.

\section{Experiments and Results}

\begin{figure*}
  \includegraphics{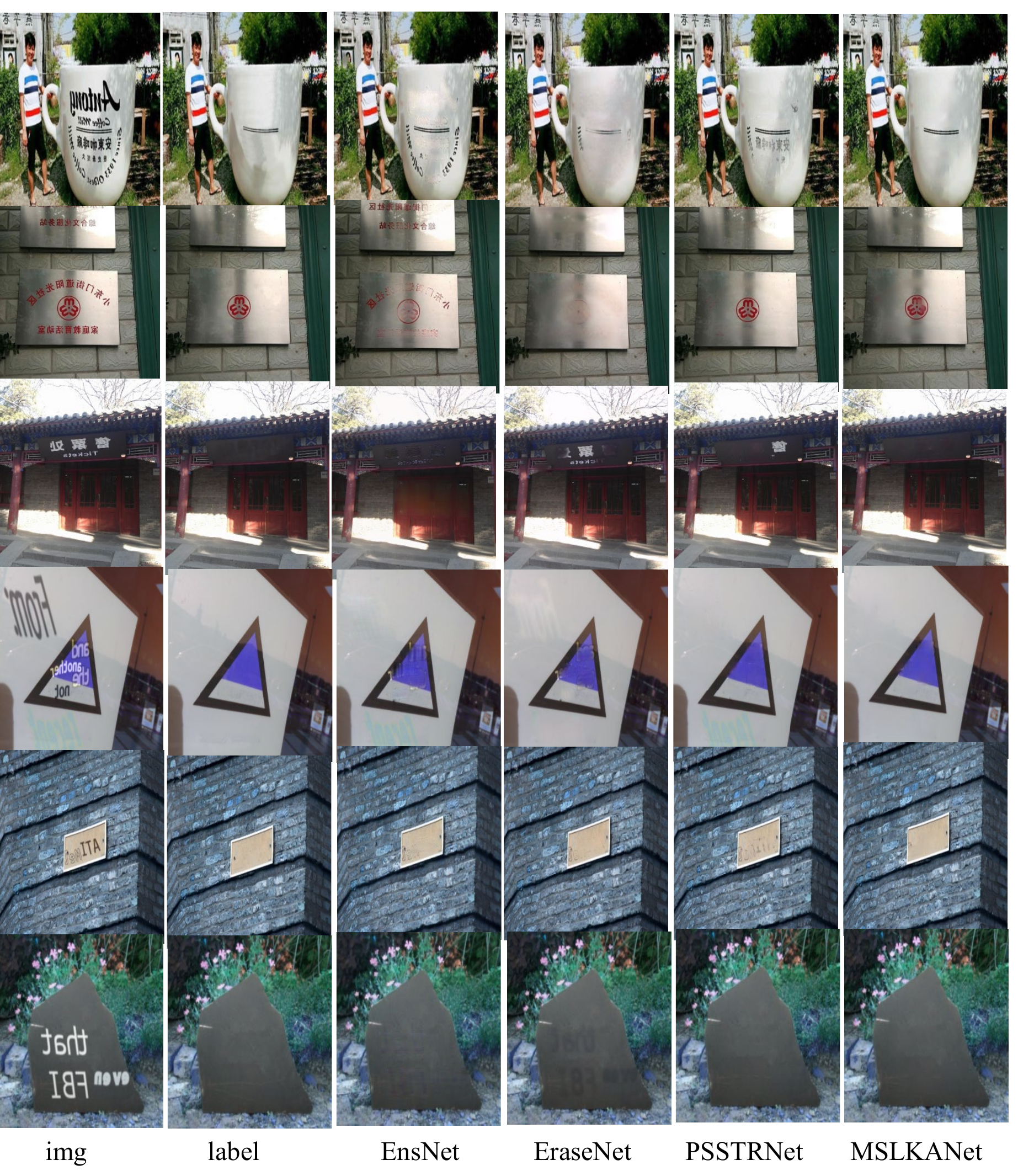}
  \caption{Comparison results with other SOTA methods on SCUT-EnsText and SCUT-Syn datasets.}
  \label{SOTA}
\end{figure*}

\subsection{Datasets and Evaluation Metrics}
\textbf{SCUT-Syn}. It is a synthetic dataset that splits 8,000 images for training and 800 images for testing. More details can be found in \cite{zhang2019ensnet}.

\textbf{SCUT-EnsText}. It contains 2,749 training images and 813 test images which are collected in real scenes. More descriptions refer to \cite{liu2020erasenet}.

\textbf{Evaluation Metrics:}. Following the previous methods \cite{zhang2019ensnet,liu2020erasenet,lyu2022psstrnet} For detecting text on the output images, we employ a text detector CRAFT\cite{baek2019character} to calculate recall and F-score. The lower, the better. PSNR, MSE, MSSIM, AGE, pEPs, and pCEPS are adopted for measuring the equality of the output images. The higher MSSIM and PSNR, and the lower AGE, pEPs, pCEPS, and MSE indicate better results.

\subsection{Implementation Details}

We train MSLKANet on the training set of SCUT-EnsText and SCUT-Syn and evaluate them on their corresponding testing sets. We follow \cite{liu2020erasenet} to apply data augmentation during the training stage. We apply a random rotation of a maximum degree of 10◦, random horizontal flip, and random vertical flip with a probability of 0.5 for data augmentation during training. The model is optimized by adamW optimizer with the weight decay of $1e^{-4}$ and the epison of $1e^{-3}$. The initial learning rate is set to be $1e^{-3}$. We also use the warmup strategy \cite{goyal2017accurate} that increases the learning rate from 0 to the initial learning rate linearly to overcome early optimization difficulties. After the learning rate warmup, we typically apply the cosine annealing strategy \cite{loshchilov2016sgdr} to steadily decrease the value from the initial learning rate. The MSLKANet is trained by a single NVIDIA GPU 3090 with a batch size of 16 and the input image size of 256 $\times$ 256.

\begin{table}
\center
  \caption{Ablation study results of different modules' effect on SCUT-Text.}
  \resizebox{\linewidth}{!}{
  \begin{tabular}{|cc|c|c|c|c|c|c|}
    \hline
    \multicolumn{2}{|c|}{Method}&PSNR&MSSIM&MSE&AGE&pEPs&pCEPs\\
    \hline
    \multicolumn{2}{|c|}{Baseline} & 33.87 & 96.42 & 0.15 & 2.4561 & 0.0171 & 0.0112 \\
    \hline
    \multicolumn{2}{|c|}{+MSLKA} & 34.89 & 97.10 & 0.0013 & 1.8228 & 0.0142 & 0.0085 \\
    \hline
    \multicolumn{2}{|c|}{+MSLKA+ASPP} & 35.07 & 97.18 & 0.0013 & 1.7527 & 0.0133 & 0.082 \\
    \hline
    \multicolumn{2}{|c|}{+MSLKA+LKSPP} & \textbf{35.27} & \textbf{97.27}  & \textbf{0.0012} & \textbf{1.6647} & \textbf{0.0129} & \textbf{0.0076} \\
    \hline
   \end{tabular}
}
\label{tab_abla}
\end{table}

\subsection{Ablation Study}

\begin{table*}
\center
  \caption{Comparison with SOTA methods and proposed method on SCUT-EnsText. R: Recall; P: Precision; F: F-score.}
  \begin{tabular}{|c|c|c|c|c|c|c|c|c|c|c|}
    \hline
    \multirow{2}{*}{Method}&\multicolumn{6}{c|}{Image-Eval}&\multicolumn{3}{c|}{Detection-Eval(\%)}\\\cline{2-10}
    &PSNR $\uparrow$ &MSSIM$\uparrow$ &MSE$\downarrow$ &AGE$\downarrow$ &pEPs$\downarrow$ &pCEPs$\downarrow$ &P$\downarrow$ &R$\downarrow$ &F$\downarrow$\\
    \hline
    Original Images  & - & - & - & - & - & - & 79.8 & 69.7 & 74.4 \\
    \hline
    Pix2pix & 26.75 & 88.93 & 0.0033 & 5.842 & 0.048 & 0.0172 & 71.3 & 36.5 & 48.3 \\
    \hline
    Scene Text Eraser & 20.60 & 84.11 & 0.0233 & 14.4795 & 0.1304 & 0.0868 & 52.3 & 14.1 & 22.2 \\
    \hline
    EnsNet & 29.54 & 92.74 & 0.0024 & 4.1600 & 0.2121 & 0.0544 & 68.7 & 32.8 & 44.4 \\
    \hline
    EraseNet & 31.62 & 95.05 & 0.0015 & 3.1264 & 0.0192 & 0.0110 & 54.1 & 8.0 & 14.0 \\
    \hline
    PERT & 33.25 & 96.95 & 0.0014 & 2.1833 & 0.0136 & 0.0088 & 52.7 & 2.9 & 5.4 \\
    \hline
    PSSTRNet & 34.65 & 96.75 & 0.0014 & 1.7161 & 0.0135 & \textbf{0.0074} & \textbf{47.7} & 5.1 & 9.3 \\
    \hline 
    MSLKANet(Ours) & \textbf{35.27} & \textbf{97.27}  & \textbf{0.0012} & \textbf{1.6647} & \textbf{0.0129} & 0.0076 & 51.3 & \textbf{3.8} & \textbf{7.0} \\
    \hline
\end{tabular}
\label{tab_sota_enstext}
\end{table*}

\begin{table*}
\center
  \caption{Comparison with SOTA methods and proposed method on SCUT-Syn.}
  \begin{tabular}{|c|c|c|c|c|c|c|c|c|}
    \hline
    Method &PSNR $\uparrow$ &MSSIM$\uparrow$ &MSE$\downarrow$ &AGE$\downarrow$ &pEPs$\downarrow$ &pCEPs$\downarrow$&Parameters$\downarrow$\\
    \hline
    Pix2pix & 25.16 & 87.63 & 0.0038 & 6.8725 & 0.0664 & 0.0300 & 54.4M  \\
    \hline
    Scene Text Eraser & 24.02 & 89.49 & 0.0123 & 10.0018 & 0.0728 & 0.0464 & 89.16M  \\
    \hline
    EnsNet & 37.36 & 96.44 & 0.0021 & 1.73 & 0.0276 & 0.0080 & 12.4M \\
    \hline
    EraseNet & 38.31 & 97.68 & 0.0002 & 1.5982 & 0.0048 & 0.0009& 19.74M  \\
    \hline
    PERT & 39.40 & 97.87 & 0.0002 & 1.4149 & 0.0045 & 0.0006 & 14.00M \\
    \hline
    PSSTRNet & 39.25 & 98.15 & 0.0002 & 1.2035 & 0.0043 & 0.0008 & \textbf{4.88M}  \\
    \hline
    MSLKANet & \textbf{41.40} & \textbf{98.58} & \textbf{0.0001} & \textbf{0.9976} & \textbf{0.0026} & \textbf{0.0003} & 9.62M \\
    \hline
    
\end{tabular}
\label{tab_sota_syn}
\end{table*}

In this section, we verify the contributions of different components MSLKA and LKSPP  on the SCUT-Text dataset. Our baseline model is implemented by a UNet-Style model similar to the pioneer method. In total, we conduct four experiments: 1. baseline, 2.baseline+MSLKA, 3.baseline+MSLKA+ASPP 4.baseline+MSLKA+LKSPP. All experiments use the same training and test settings. 

\textbf{MSLKA}. MSLKA aims to obtain long-range dependencies between the text regions and the backgrounds at various granularity levels. As shown in table\ref{tab_abla}, our MSLKA provides significant enrichments over the baseline on all metrics. 

\textbf{LKSPP}. LKSPP is proposed to perceive more valid pixels in the spatial dimension while maintaining large receptive fields and lowering the cost of computation. It can also provide global information According to the results shown in table\ref{tab_abla}, our LKSPP is helpful in both text removal and background in painting.

\subsection{Comparison with State-of-the-Art Approaches}
In this section, we compare our proposed MSLKANet with six state-of-the-art methods: Pix2pix\cite{isola2017image}, Scene Text Eraser\cite{nakamura2017scene}, EnsNet\cite{zhang2019ensnet}, EraseNet\cite{liu2020erasenet}, PERT \cite{wang2021pert} and PSSTRNet \cite{lyu2022psstrnet}, on both SCUT-EnsText and SCUT-Syn datasets. Qualitative and quantitative results are illustrated in Fig.\ref{SOTA}, table\ref{tab_sota_enstext} and \ref{tab_sota_syn}, respectively.

\textbf{Qualitative Comparison}. As shown in the 1st, 3rd, and 4th rows of Fig.\ref{SOTA}, our model can remove multi-scale texts, especially for the larger scale text.  Compared with other state-of-the-art methods, the results of our proposed MSLKANet have significantly fewer color discrepancies and blurriness, and cleaner background. They demonstrate our model could generate more natural results on text removal and background inpainting results.

\textbf{Quantitative Comparison}. As shown in table \ref{tab_sota_syn} and \ref{tab_sota_syn}, our method produces the best scores on most scene text removal and image inpainting equality evaluation metrics for both SCUT-EnsText and SCUT-Syn datasets. It indicates the advances of our proposed MSLKANet.

\subsection{Generalization to general object removal}
\begin{figure}
  \includegraphics[width=\linewidth,height=0.7\linewidth]{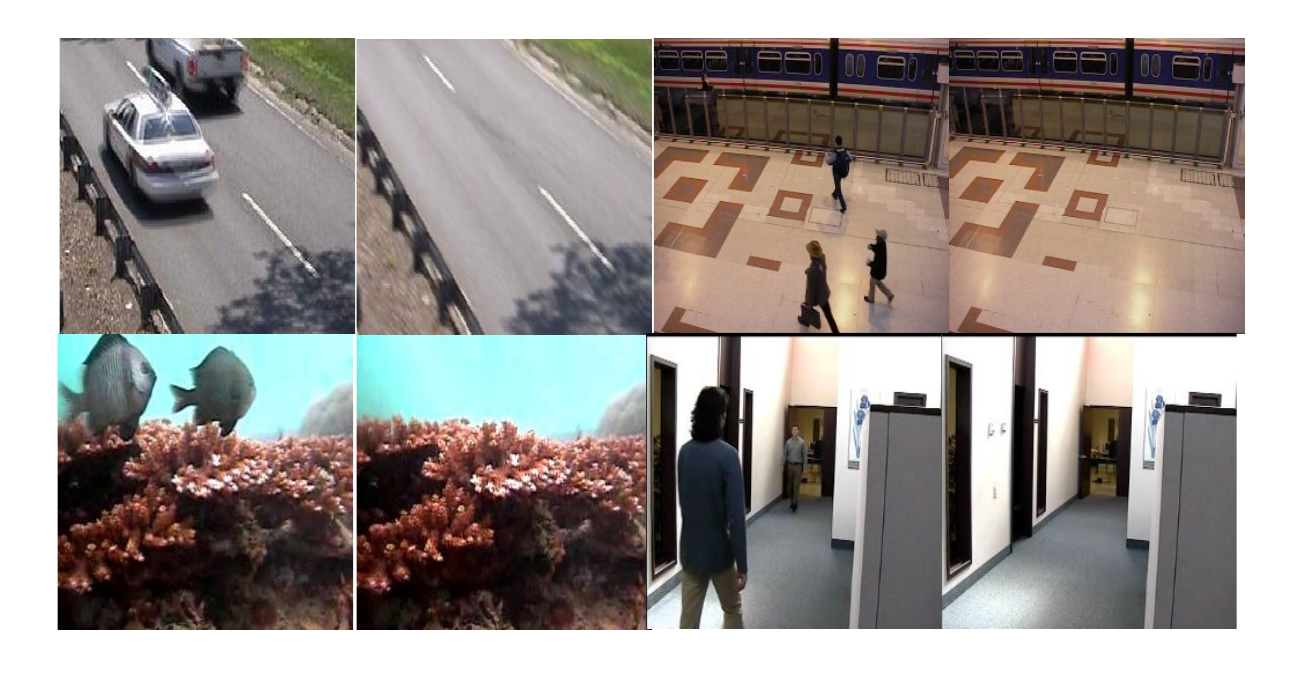}
  \caption{The results on several sub-datasets of SBMnet dataset for general object removal.}
  \label{generation}
\end{figure}

Following the important scene text removal network EnsNet\cite{zhang2019ensnet}, we also execute an experiment on the general object removal task to test the generalization ability of MSLKANet. We also evaluate MSLKANet on the well-known Scene Background Modeling (SBMnet) dataset \cite{jodoin2017extensive}, which is proposed for background estimation algorithms. The qualitative results are shown in Fig.\ref{generation}, our MSLKANet reconstructs a natural background. This means that our method can be generalized to remove pedestrians, cars, and fish without specific adjustments. It can certify the generalization of our proposed  MSLKANet.

\section{limitation}
\begin{figure}
  \includegraphics[width=\linewidth,height=0.6\linewidth]{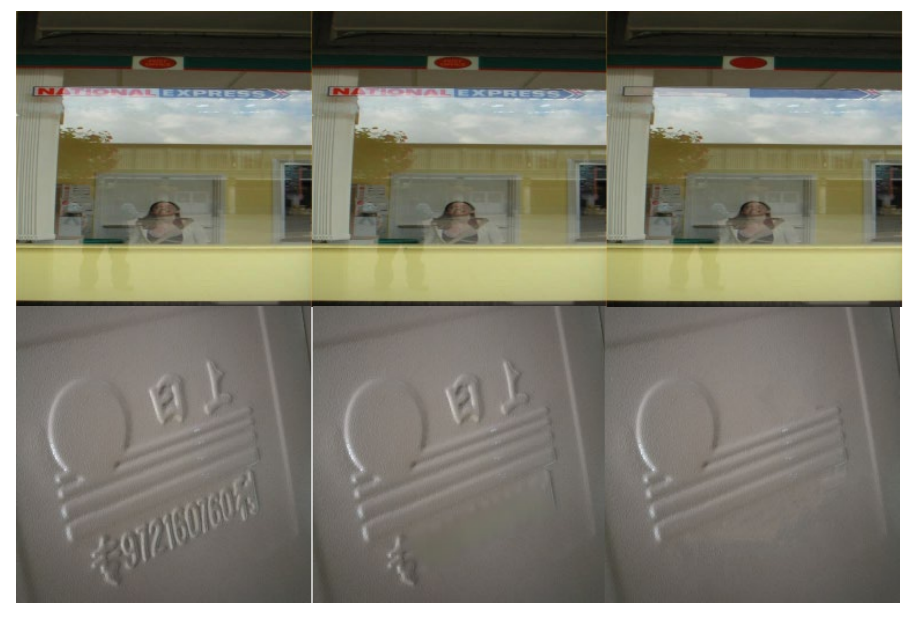}
  \caption{The failure cases of MSLKANet on SCUT-EnsText. From left to right: input, output, and label.}
  \label{fig_failure_case}
\end{figure}
As shown in the 1st row of Fig.\ref{SOTA} and Fig.\ref{fig_failure_case}, MSLKANet fails in handling texts blended into the environment and there are still some texts that are not be removed. Besides, there is much room for improvement in erasing the specular text.

In addition, it is difficult to insert the extra block into the existing methods due to the old and disordered architecture. So, we will rebuild the all the methods following the modern network architecture design \cite{ronneberger2015u,liu2022convnet,he2016deep} and evaluate them in fair settings.

\section{Conclusion}
In this paper, we present a single-stage multi-scale network MSLKANet for scene text removal in full images. In experiments, we find that the pioneer method can get significant improvements by only changing training data from the cropped image to the full image. For obtaining large perceptive fields and global information, we propose multi-scale large kernel attention (MSLKA) to obtain long-range dependencies between the text regions and the backgrounds at various granularity levels. Furthermore, we combine the large kernel decomposition mechanism and atrous spatial pyramid pooling to build large kernel spatial pyramid pooling (LKSPP), which can fuse more valid pixels in the spatial dimension while maintaining large receptive fields and low cost of computation. Extensive experimental results indicate that the proposed method achieves state-of-the-art performance on both synthetic and real-world datasets and the effectiveness of the proposed components MSLKA and LKSPP.

{\small
\bibliographystyle{ieee_fullname}
\bibliography{egbib}
}

\end{document}